%% file: main.tex
\documentclass[final]{siamltex}
\usepackage{algorithm,algpseudocode}
\usepackage{latexsym, graphicx, epsfig, amsmath, amsfonts,amssymb,bm}
\usepackage{caption, subcaption}
\usepackage{epstopdf}
\usepackage{booktabs}
\usepackage{array}
\usepackage{color}
\usepackage{url}
\usepackage{comment}
\usepackage{diagbox}

\def\Dt{\Delta t}

\allowdisplaybreaks

\def\be{\begin{equation}}
\def\ee{\end{equation}}

\def\x{\mathbf{x}}
\def\y{\mathbf{y}}
\def\f{\mathbf{f}}

\def\z{\mathbf{z}}

\def\N{\mathbf{N}}
\def\Nc{\mathbf{N}}

\def\PPh{\boldsymbol{\Phi}}
\def\PPs{\boldsymbol{\Psi}}
\def\nnu{\boldsymbol{\nu}}

\def\SSigma{\boldsymbol{\Sigma}}

\def\E{{\mathbb E}}
\def\R{{\mathbb R}}

\def\I{\mathbf I}

\def\A{\mathbf{A}}
\def\D{\mathcal{D}}

\def\u{\mathbf{u}}
\def\v{\mathbf{v}}
\def\V{\mathbf{V}}
\def\w{\mathbf{w}}
\def\U{\mathbf{U}}

\def\0{\mathbf{0}}

\newcommand{\argmin}{\operatornamewithlimits{argmin}}

\title{Robust Modeling of Unknown Dynamical Systems via Ensemble Averaged Learning}

\author{Victor Churchill\footnotemark[1]\and Steve Manns\thanks{Department of Mathematics,
		The Ohio State University, Columbus, OH 43210, USA. Emails:
		{\tt churchill.77@osu.edu, manns.79@osu.edu, xiu.16@osu.edu} Funding: This 
		work was partially supported by AFOSR FA9550-22-1-0011.}\and Zhen
		Chen\thanks{Department of Mathematics, Dartmouth College,
		Email: {\tt zhen.chen@dartmouth.edu}}\and Dongbin Xiu\footnotemark[1]
				}

\begin{document}
\maketitle
\begin{abstract}
Recent work has focused on data-driven learning of the evolution of unknown systems via deep neural networks (DNNs), with
the goal of conducting long time prediction of the evolution of the unknown system. Training a DNN with low generalization error is a particularly important task in this case as error is accumulated over time. Because of the inherent randomness in DNN training, chiefly in stochastic optimization, there is uncertainty in the resulting prediction, and therefore in the generalization error. Hence, the generalization error can be viewed as a random variable with some probability distribution. Well-trained DNNs, particularly those with many hyperparameters, typically result in probability distributions for generalization error with low bias but high variance. High variance causes variability and unpredictably in the results of a trained DNN. This paper presents a computational technique which decreases the variance of the generalization error, thereby improving the reliability of the DNN model to generalize consistently. In the proposed ensemble averaging method, multiple models are independently trained and model predictions are averaged at each time step. A mathematical foundation for the method is presented, including results regarding the distribution of the local truncation error. In addition, three time-dependent differential equation problems are considered as numerical examples, demonstrating the effectiveness of the method to decrease variance of DNN predictions generally.
\end{abstract}

\input Introduction
\input Setup
\input Method

\input Examples

\input Conclusion

\bibliographystyle{siamplain}
\bibliography{neural,LearningEqs,ensemble}

\end{document}

%% file: Introduction.tex
\section{Introduction} \label{sec:intro}

Learning unknown dynamical systems, including both ordinary and partial differential equations, from data has been a prominent area of research in recent years. One way to approach this problem is to construct a mapping from the state variables to their time derivatives, which can be achieved via sparse approximation from a large dictionary set. In certain circumstances, exact equation recovery is possible. See, for example, \cite{brunton2016discovering} and its many extensions in recovering both ODEs (\cite{brunton2016discovering,kang2019ident, schaeffer2017sparse,schaeffer2017extracting, tran2017exact}) and PDEs (\cite{rudy2017data, schaeffer2017learning}). Deep neural networks (DNNs) have also been used to construct this mapping. See, for example, ODE modeling in \cite{lu2021deepxde,qin2018data,raissi2018multistep,rudy2018deep}, and PDE modeling in \cite{long2018pde,long2017pde,lu2021learning,raissi2017physics1,raissi2017physics2,raissi2018deep,sun2019neupde}. Recently, in similar spirit to this paper, ensembles of models have been used to improve robustness \cite{fasel2021ensemble}.

Another approach for learning unknown systems, which is the focus in this paper, is to construct a mapping between two system states separated by a short time \cite{qin2018data}. Unlike the previous approach, constructing this flow map does not directly formulate the underlying equations. Instead, if trained properly such that the true flow map is accurately approximated, it allows for the definition of an accurate predictive model for evolution of the unknown system. One advantage of the flow map based approach is that it does not require temporal derivative data, which can be difficult to acquire or subject to large errors when approximated. This approach relies on DNNs, in particular residual networks (ResNet \cite{he2016deep}) to approximate the flow map. Introduced to model autonomous systems in \cite{qin2018data}, flow map based DNN learning of unknown systems has been extended to model non-autonomous systems \cite{QinChenJakemanXiu_SISC},  parametric dynamical systems
\cite{QinChenJakemanXiu_IJUQ}, partially observed dynamical systems
\cite{FuChangXiu_JMLMC20}, as well as PDEs in modal space
\cite{WuXiu_modalPDE}, and nodal space \cite{chen2022deep}.

In the flow map based approach to DNN modeling, networks are trained on training data sets by minimizing some particular loss function. The vast majority of the algorithms used to minimize these loss functions are randomized optimization algorithms, with stochastic gradient descent (SGD) and adaptive moment estimation (Adam \cite{Adam_2014}) widely used in many DNN packages.
The randomness in these algorithms results in variability in the training results, and therefore in prediction accuracy as well. Consequently, even with the same training data set and the same choice of the
training algorithm, the resulting trained DNN models differ from each other when one ``re-trains'' the network. 
The differences in the trained DNNs, albeit small sometimes, result in performance differences in the models. It is not uncommon for one to obtain some ``good''
DNN models and some ``bad'' models, simply because the models are trained at a different time, or on different servers, when everything else remains the same. This occurs because the execution of the stochastic optimization relies on fixing a random seed at the start of the algorithm.  Each time the algorithm is started, the seed is reset (by default) in a random number generator and subsequently changes the entire progression of stochastic optimization algorithms.
The differences in training results become much more prominent for modeling unknown dynamical systems. For system predictions, the baseline DNN model
needs to be executed repetitively for long-term system predictions. Therefore, any differences in the DNN models will be amplified over time in the form of function composition and produce sometimes vastly different long-term system predictions.

In this paper, we present a robust DNN modeling approach, in the sense that the resulting DNN model possesses (much) less randomness. This is achieved
by taking an ensemble average at each time step of multiple independently trained DNNs, using the same training data set and an identical stochastic training algorithm.
The ensemble averaged DNN model has (much) smaller variance in its outputs and subsequently produces more consistent performance with less variability.
%
The concept of decreasing variance through ensemble averaging of neural network models is not new, cf. \cite{breiman1996bagging,efron1992bootstrap,horn2012large,naftaly1997optimal,perrone1993improving,perrone1994pulling,perrone1992networks,ueda1996generalization}.\footnote{Perhaps the most well-known ensemble method is bootstrap aggregating or  ``bagging'', \cite{breiman1996bagging,efron1992bootstrap}, which averages multiple low-bias learners to reduce variance. Typically, the variance due to variation in training data is reduced by dividing a limited training data set into multiple overlapping sets.}
However, the concept is typically used in applications with single-step prediction such as classification or regression \cite{perrone1992networks}.
Here we focus on DNN modeling of unknown dynamical systems, where the DNN needs to be recursively executed for long-time system predictions.
Ensemble averaging in this case is {\em not} an average of all the predictions from each individually trained model. (This will lead to erroneous and unphysical predictions.)  Instead, at each time step, all individual DNN models
start with the same input condition and produce their own one-step predictions. These individual predictions are then averaged into a single prediction at the new
time step, which is then used as the input condition for the next prediction, and so on. Our analysis demonstrates that the variance (due to randomness in training) at each time
step is asymptotically $\sim 1/K$ times of that of the individual models, where $K$ is the total number of DNN models. This results in smaller local
truncation error and consequently smaller overall numerical errors for the ensemble averaged predictive DNN model.


%% file: Setup.tex
\section{Flow Map Modeling of Unknown Dynamical Systems} \label{sec:setup}

We are interested in constructing effective models for the evolution
laws behind dynamical data.
%
Throughout this paper our discussion will be over discrete time
instances with a constant time step $\Dt$,
\be \label{tline}
t_0<t_1<\cdots, \qquad
t_{n+1} - t_n = \Dt, \quad \forall n.
\ee
Generality is not lost with the constant time step assumption. 
We will also use a subscript to denote the time variable of
a function, e.g., $\x_n = \x(t_n)$.


\subsection{ResNet Modeling}

For an unknown autonomous system,
\be\label{eq:ODE}
\frac{d\x}{dt} = \f(\x), \qquad \x\in\R^d,
\ee
where $\f:\R^d\to \R^d$ is not known. Its flow map depends only
on the time difference but not the actual time, i.e.,
$ \x_n = \PPh_{t_n-t_s}(\x_s)$. Thus, the solution over one time step
satisfies
\be
\x_{n+1} = \PPh_{\Dt}(\x_n) = \x_n + \PPs_{\Dt}(\x_n),
\ee
where $\PPs_{\Dt} = \PPh_{\Dt} - \mathbf{I}$, with $\mathbf{I}$ as the
identity operator.

When data for the state variables $\x$ over the time stencil
\eqref{tline} are available, they can be grouped into pairs separated by
one time step
$$
\{\x^{(m)}(0), ~~\x^{(m)}(\Dt)\},\quad m=1,\ldots,M,
$$
where $M$ is the total number of such data pairs. This is the training data set.
One can define a residual network (ResNet) (\cite{he2016deep}) in the
form of
\be
\mathbf{y}^{out} = \left[\mathbf{I}+\mathbf{N} \right](\mathbf{y}^{in}),
\ee
where $\N:\R^d\to\R^d$ stands for the mapping operator of a standard
feedforward fully connected neural network.
The network is then trained by using the training data set 
and minimizing the mean squared loss function
\be
\sum_{m=1}^M \left\| \x^{(m)}(\Dt) - (\I+\N)(\x^{(m)}(0))\right\|^2.
\ee
The trained network thus accomplishes
$$
\x^{(m)}(\Dt) \approx \x^{(m)}(0)+\N(\x^{(m)}(0)), \qquad \forall m.
$$
Once the network is trained to satisfactory accuracy, it can be used as a predictive model
\be \label{ResNet}
\x_{k+1} =  \x_k + \N(\x_k), \qquad k=0,1,\dots,
\ee
for any initial condition $\mathbf{x}(t_0)$. This framework was
proposed in  \cite{qin2018data}, with extensions to parametric systems
and time-dependent (non-autonomous) systems (\cite{QinChenJakemanXiu_IJUQ,QinChenJakemanXiu_SISC}).

\subsection{Memory based DNN Modeling}

A notable extension of the flow map modeling in \cite{qin2018data} is
for systems with missing variables. Let $\x = (\z; \w)$, where
$\z\in\R^n$ and $\w\in\R^{d-n}$. Let $\z$ be the observables and $\w$
be the missing variables. That is, no information about or data from $\w$ are
available. When data are available for $\z$, it is possible
to derive a system of equations for $\z$ only. The celebrated
Mori-Zwanzig formulation  (\cite{mori1965, zwanzig1973}) asserts that the system for $\z$ requires a
memory integral. By making a mild assumption that the memory is of
finite length (which is problem dependent), memory-based DNN
structures were investigated \cite{WangRH_2020, FuChangXiu_JMLMC20}.
While \cite{WangRH_2020} utilized LSTM (long short-term memory) networks,
\cite{FuChangXiu_JMLMC20} proposed a fairly simple DNN structure, in
direct correspondence to the Mori-Zwanzig formulation,
that takes the following mathematical form,
\be
      \z_{n+1} = \z_n +
      \N(\z_n,\z_{n-1}, \dots,\z_{n-n_M}),  \qquad n\geq n_M,
      \ee
      where $n_M\geq 0$ is the number of memory term in the model. In
      this case, the DNN operator is $\N:\R^{d\times (n_M+1)}\to
      \R^d$. The special case of $n_M=0$ corresponds to the standard
      ResNet model \eqref{ResNet}, which is for modeling systems
      without missing variables (thus no need for memory).

\subsection{Modeling of PDE}\label{sec:PDE}

The flow map learning approach has also been applied in modeling PDE. When the
solution of the PDE can be expressed using a fixed basis, the learning
can be conducted in modal space. In this case, the ResNet approach can
be adopted. See \cite{WuXiu_modalPDE} for details.
When data of the PDE solutions are available as nodal values over a
set of grids in physical space, its DNN learning is more involved. In this
case, a DNN structure was developed in \cite{chen2022deep}, which can
accommodate the situation when the data are on unstructured grids. It
consists of a set of specialized layers including disassembly layers and an
assembly layer, which are used to model the potential differential
operators involved in the unknown PDE.
The proposed DNN model defines the following mapping,
\be \label{NN_simple}
\w_{n+1} = \w_n + \A(\N_1(\w_n), \dots, \N_J(\w_n)),
\ee
where $\N_1,\dots, \N_J$ are the operators for the disassembly layers
and $\A$ the operator for the assembly layer. The DNN modeling
approach was shown to be highly flexible and accurate to learn a
variety of PDEs (\cite{chen2022deep}).

%% file: Method.tex
\section{Ensemble Averaged Learning} \label{sec:method}

We now present the ensemble average modeling of unknown systems. We
first discuss ensemble averaging for DNN training in a general
setting, and then its application to modeling unknown dynamical
systems and the associated numerical properties such as error bounds.

\subsection{Ensemble Averaged DNN Training} \label{sec:average}

Let us consider an unknown map
$
\f: \R^\ell\to\R^d.
$
Let
\be
D = \{ \x^{(m)}, \y^{(m)}\}_{m=1}^M
\ee
be a training data set, where $\y^{(m)} = \f(\x^{(m)})$ and may also
contain noise.
Let $\Nc (\cdot; \Theta):\R^\ell\to\R^d$ be a DNN map with a fixed architecture that approximates
$\f$, where $\Theta$ denotes the set of hyper-parameters in the
network. These parameters include the weights and biases for all neurons.
 The parameters are determined
by network training, which is minimizing a loss function over
the training data set $D$,
\be
\Theta^* = \argmin_{\Theta} \mathcal{L}(\Theta; D).
\ee
Different loss functions are used for different problems. In flow map
learning of dynamical systems, the main topic of this paper, mean squared
loss is typically used, 
\be
\mathcal{L}(\Theta; D) =
\frac{1}{M}\sum_{m=1}^M\left\|\y^{(m)} - \Nc(\x^{(m)};\Theta)\right\|^2.
\ee
We remark that the discussion here is rather independent of the choice
of the loss function.

In practice, the network training is carried out by an iterative
algorithm. Hereafter we focus exclusively on stochastic optimization
algorithms. The most widely used and representative examples include
Adam \cite{Adam_2014} and certain versions of stochastic gradient decent
(SGD). Let $\mathcal{S}$ be the algorithm set, which includes the
chosen algorithm, e.g., Adam or SGD, along with its associated user 
choices including
initial choice of the parameter $\Theta$, stopping criterion for the iteration, size of
mini-batch, number of epochs, etc. An important feature of the stochastic optimization
is its inherent randomness after choosing $\mathcal{S}$. The most notable sources of
randomness include:
\begin{itemize}
  \item Initial condition for $\Theta$. Users choose the distribution of
    the initial condition for the optimization. The optimization algorithm then
    samples from the distribution for the initial condition.
    \item Mini-batch. When mini-batches are used (which is often the
      case), they are usually randomly selected and shuffled after
      each epoch. This affects the step direction in the optimization.
    \end{itemize}

    Due to the inherent randomness of the optimization, the parameter
    $\Theta$ obtained at the end of the stochastic optimization
    algorithm follows some probability distribution,
    \be
    \Theta \sim \D_\Theta(D, \mathcal{S}),
    \ee
where $\D_\Theta$ depends on the training data set $D$ and the
optimization algorithm setting $\mathcal{S}$. Hereafter we
shall assume $D$ and $\mathcal{S}$ are fixed and suppress
their dependence in the expression, unless confusion arises otherwise.
Subsequently,
the trained network $\Nc$, when evaluated at any input $\x$, also
follows a distribution,
\be 
\Nc|\x \sim \#_{\Nc(\x)} \D_\Theta,
\ee
where $\#_{\Nc(\x)}$ stands for the push forward measure by the mapping
  $\Nc(\x)$.

Let us assume that, with a chosen optimization algorithm 
along with its fixed settings $\mathcal{S}$, one conducts $K\geq 1$ 
independent trainings for the given training data set $D$. 
In practice, independent trainings can be conducted by changing the random seed.
Let $\Theta^{(i)}$, $i=1,\dots,K$, be the resulting network parameters
obtained by the independent trainings. We then have $\Theta^{(i)}$ are i.i.d.,
i.e., for $i=1,\dots, K,$
\be \label{Th_distr}
\Theta^{(i)} \quad\textrm{i.i.d. } \sim \D_\Theta.
\ee

Let $\Nc^{(i)} = \Nc(\cdot; \Theta^{(i)})$, $i=1,\dots,K$, be the
resulting DNN network operators from the $K\geq 1$ independent
training. They are also independent draws from the push forward
measure defined by the network structure, i.e.,
\be \label{N_distr}
\Nc^{(i)} |\x \quad \textrm{i.i.d. }\sim \#_{\Nc(\x)}\D_\Theta, \qquad i=1,\dots,K.
\ee
Let
    \be \label{mean}
  \nnu(\x) = \E\left[\Nc(\x;\Theta)\right],
  \qquad
  \SSigma(\x) = \text{Cov}(\Nc(\x;\Theta))
  \ee
  be  the mean and covariance.

In ensemble averaged learning, we define the final DNN mapping
operator as the average of the $K\geq 1$
independent trained network operators,
\be \label{NA}
\Nc^{(A)}(\cdot) = \frac{1}{K}\sum_{i=1}^K \Nc^{(i)}(\cdot).
\ee

The following statement follows immediately from the celebrated
central limit theorem.
\begin{proposition} \label{prop}
  For a given training data set $D$ and a chosen stochastic
  optimization training algorithm with chosen settings
  $\mathcal{S}$, let $\Theta^{(i)}$, $i=1,\dots, K$, be the neural
  network parameter sets obtained from $K\geq 1$ independent training, and
  $\Nc^{(i)}=\Nc(\cdot; \Theta^{(i)})$  the corresponding neural network
  mappings.
  Then, the ensemble averaged neural network mapping \eqref{NA}
  then follows,
  \be
  \sqrt{K} \left( \Nc^{(A)}|\x - \nnu(\x)\right) \to
  \mathcal{N}_d(\0,\SSigma),  \qquad K\to\infty,
  \ee
  where
  $\nnu(\x)$ and $\SSigma(\x)$ are defined in \eqref{mean} and 
$\mathcal{N}_d(\0,\SSigma)$ is the $d$-variate normal distribution with
  zero mean and covariance $\SSigma$ and the convergence is in distribution.
  \end{proposition}

\subsection{Ensemble Averaged Modeling for Dynamical  Systems}

Applying ensemble averaging to unknown system modeling is
straightforward. It requires independent training of several DNN
models using the same training data set, algorithm, and algorithm settings. During the system
prediction, each DNN model is executed in parallel over a single
prediction time step and then the results are averaged as the
prediction result. This averaged prediction result is used as the next initial condition for predicting the next time step, and so on.

For a detailed discussion, we use the following generic formulation
to incorporate the various forms of DNN
modeling of unknown systems reviewed in Section \ref{sec:setup}.
Let $n_M\geq 0$ be the number of memory terms and
$$
\V_{n} = (\v_n; \v_{n-1};\cdots, \v_{n-n_M})
$$
a concatenated vector.
We can write the various DNN models in Section \ref{sec:setup} as
\be\label{eq:model}
  \v_{n+1} = \v_n + \Nc(\V_{n}; \Theta), \qquad n\geq n_M,
  \ee
  where the ``zero-memory'' case of $n_M=0$ corresponds to either the ResNet model \eqref{ResNet} or the PDE model \eqref{NN_simple}. 
The DNN operator defines a mapping
$\mathbf{N}:\mathbb{R}^{d\times
  (n_M+1)}\rightarrow\mathbb{R}^{d}$ with a hyperparameter set $\Theta$.
Let $D$ be the training dataset, consisting of data of the unknown
system over sequences of $(n_M+2)$ steps. The DNN model can be trained
by using the first $(n_M+1)$ steps of the data as the inputs and the last
step of the data as the outputs. Upon choosing a training algorithm,
Adam or SGD, and fixing its associated training settings $\mathcal{S}$,
we then conduct $K\geq 1$ independent training over the same dataset
$D$. In practice, this can be readily achieved by manually changing the random seed.

Let $\Nc^{(i)} = \Nc(\cdot; \Theta^{(i)})$, $i=1,\dots,K$, be the
$K\ge1$ independently trained DNN mappings. Then, each individual DNN
model is
\be\label{Vi}
  \v^{(i)}_{n+1} = \v_n^{(i)} + \Nc(\V^{(i)}_{n}; \Theta), \qquad n\geq n_M,
  \ee
  with initial conditions
  $$
\V_{n_M}^{(i)} = (\u_{n_M}; \u_{n_M-1}, \cdots, \u_0), \qquad
\forall i=1,\dots,K,
$$
where $\u_n = \u(t_n)$ are the exact solutions of the unknown system.

In ensemble averaged DNN model
prediction, let
$$
\V^{(A)}_{n} = (\v_n^{(A)}; \v_{n-1}^{(A)};\cdots,
\v_{n-n_M}^{(A)}), \qquad n\geq n_M.
$$
With the true initial conditions 
$$
\V_{n_M}^{(A)} = (\u_{n_M}; \u_{n_M-1}, \cdots, \u_0),
$$
the ensemble averaged model consists of the following steps:
for $n\geq n_M$,
\be \label{VA}
\left\{
\begin{split}
  &\v_{n+1}^{(i)} = \v_n^{(A)} + \Nc(\V_{n}^{(A)}; \Theta^{(i)}), \qquad
  i=1,\dots, K,\\
  &\v_{n+1}^{(A)} = \frac{1}{K} \sum_{i=1}^K \v_{n+1}^{(i)}.
  \end{split}
\right.
\ee
Or, equivalently, this can be written as
  \be \label{vA}
  \v_{n+1}^{(A)} = \v_n^{(A)} +  \Nc^{(A)}(\V_{n}^{(A)})\qquad n\geq n_M,
  \ee
  where
  \be \label{NA0}
  \Nc^{(A)}(\cdot) := 
  \frac{1}{K}
  \sum_{i=1}^K \Nc(\cdot; \Theta^{(i)}).
  \ee
  
  Note that in ensemble averaged DNN prediction, all independent DNN
  models use exactly the same initial conditions at each time
  step. The model predictions from the individual models are 
  immediately averaged after marching forward one time step. The
  averaged prediction is then used as the new initial condition for
  all of the individual models for the next time step. This is
  fundamentally different from the naive averaging the prediction
  results (over a certain time horizon) of each individual models.

  \subsection{Properties}

  To discuss the errors of the proposed ensemble
  averaged DNN modeling approach, we consider an
  ``intermediate'' quantity, the one-step network prediction starting
  from the true initial condition.

Let $\u_n = \u(t_n)$ be the exact solutions of the unknown system, and 
for $n_M\geq 0$, 
  $$
\U_{n} = (\u_n;\cdots, \u_{n-n_M}), \qquad n\geq n_M.
$$
We then define, for all $i=1,\dots, K$,
\be \label{w_i}
\w_{n+1}^{(i)} = \u_n + \Nc(\U_{n}; \Theta^{(i)})
.
\ee
These are the one-step predictions of the $K$ independently training
DNN models from the exact solution states.

We then have, based on the derivation in Section \ref{sec:average},
$$
\w_{n+1}^{(i)}\qquad \textrm{i.i.d. } \sim  \#_{\Nc(\U_{n})} \D_\Theta, 
$$
with 
   \be \label{nu_n}
  \nnu_{n+1} = \E\left[\Nc(\U_{n};\Theta)\right],
  \qquad
  \SSigma_{n+1} = \textrm{Cov}(\Nc(\U_{n};\Theta))
  \ee
  as the mean and covariance, respectively.
Similarly, let
\be \label{w_a}
\w_{n+1}^{(A)}   = \u_n + \Nc^{(A)}(\U_{n})
,
\ee
where $\Nc^{(A)}$ is the averaged DNN operator defined in \eqref{NA0}.
By Proposition \ref{prop}, we immediately have
  \be \label{local_WA}
  \sqrt{K} \left( \w_{n+1}^{(A)} - \nnu_{n+1}\right) \to
  \mathcal{N}_d(\0,\SSigma_{n+1}),  \qquad K\to\infty.
  \ee
Note that while here only one-step prediction from true initial
conditions is considered, the convergence to a Gaussian distribution
holds at all time steps $t_n$ since ensemble averaging is performed at
each time step in prediction. 

To discuss the numerical errors of the proposed ensemble averaged DNN
modeling, we focus exclusively on scalar systems for notational convenience.
 This is done without loss of
    generality, because the numerical errors for vector systems are merely
    combinations of the 
    component-wise errors. For example, in examining the error in
    $\w_{n+1}$ against $\u_{n+1}$, which is the topic of local
    truncation error in the next section, it suffices to examine the
    errors in each components of $\w_{n+1}$. We shall also use
    $\|\cdot\|_\omega$ to denote the $L^2_\omega$ norm in the
    probability space. For example, in the scalar version of \eqref{w_i},
    \be
    \left\| w_{n+1}^{(i)}\right\|_\omega := \left(\E\left[\left(
        w_{n+1}^{(i)}\right)^2\right]\right)^{1/2},
    \ee
    where the expectation is with respect to the probability measure
    induced by $\D_\Theta$.

  \subsubsection{Local Truncation Error}

Let
$u_n = u(t_n)$ be the exact solutions and 
for $n_M\geq 0$,
  $$
U_{n} = (u_n; u_{n-1};\cdots, u_{n-n_M}), \qquad n\geq n_M.
$$
For each $i=1,\dots,K$, independently trained DNN model $N^{(i)} = N(\cdot, \Theta^{(i)})$,
we have
\be
u_{n+1} = u_n + N^{(i)}(U_{n})+ \Dt \cdot \tau_{n+1}^{(i)}, \qquad
i=1,\dots, K.
\ee
where $\tau_{n+1}^{(i)}$ defines the local truncation error for the
$i$-th DNN model at time $t_{n+1}$.
This formulation follows the standard nomenclature of numerical analysis for
time integrator schemes.
From \eqref{w_i},
\be
w_{n+1}^{(i)} = u_n + N^{(i)}(U_{n}), \qquad i=1,\dots, K,
\ee
the mean squared local truncation error follows
\be\label{mse1}
\begin{split}
\left\|\Dt \cdot \tau_{n+1}^{(i)}\right\|_\omega^2 &= \E\left[ \left(w_{n+1}^{(i)} -
  u_{n+1}\right)^2\right] \\
&= \E\left[ \left(w_{n+1}^{(i)} -
    \E[w_{n+1}^{(i)}]\right)^2\right] + \E\left[ \left(\E[w_{n+1}^{(i)}] - u_{n+1}\right)^2\right] \\
& = \textrm{Var}\left(w_{n+1}^{(i)}\right) + \left(\nu_{n+1}- u_{n+1}\right)^2 \\
& = \textrm{Var}\left(w_{n+1}^{(i)}\right) + \textrm{Bias}\left(w_{n+1}^{(i)}\right)^2,
\end{split}
\ee
where $\nu_{n+1} = \E[w_{n+1}^{(i)}]$ is the mean and $\textrm{Var}\left(w_{n+1}^{(i)}\right)$
the variance, as the scalar counterparts of the mean and covariance
defined in \eqref{nu_n}. The bias is the difference between the mean
prediction and the true solution. We see that the mean squared local truncation error can thus be divided into variance and squared bias.

Similarly, the local truncation error of the averaged model $N^{(A)}$
can be defined via the following:
\be
u_{n+1} = u_n + N^{(A)}(U_{n})+ \Dt\cdot \tau_{n+1}^{(A)}.
\ee
The mean squared local truncation error follows
\be\label{mse2}
\begin{split}
\left\|\Dt \cdot \tau_{n+1}^{(A)}\right\|_\omega^2 &= \E\left[ \left(w_{n+1}^{(A)} -
  u_{n+1}\right)^2\right] \\
& = \textrm{Var}\left(w_{n+1}^{(A)}\right) + \left(\E[w_{n+1}^{(A)}]- u_{n+1}\right)^2 \\
& = \textrm{Var}\left(w_{n+1}^{(A)}\right) + \textrm{Bias}\left(w_{n+1}^{(A)}\right)^2.
\end{split}
\ee

The result of \eqref{local_WA} indicates that, for sufficiently large $K$, $w_{n+1}^{(A)}$ follows
a Gaussian distribution 
with mean $\E[w_{n+1}^{(A)}]=\nu_{n+1}$, which is same as the mean of the
individual models $w_{n+1}^{(i)}$, and variance $\sim 1/K$ of the
variance of the individual models. Consequently, the ensemble averaged
model has a smaller local truncation error, in the sense of
\be \label{better_lte}
\left\|
  \tau_{n+1}^{(A)}\right\|_\omega < \left\|\tau_{n+1}^{(i)}\right\|_\omega, \qquad
\forall i=1,\dots, K.
\ee

While the variance can be
decreased by increasing the number of DNN ensembles $K$, the bias is dependent on the fixed network
architecture, the training dataset $D$, the training optimization
algorithm setting $\mathcal{S}$. In practice, a
particularly effective way to reduce bias is to increase the number of
training epochs to achieve smaller training errors.

\subsubsection{Error Bound}

To quantify the overall numerical error and establish convergence, we make a
rather mild assumption that the trained DNN operator is Liptschitz
continuous with respect to its inputs. More specifically, let $n_M\geq
0$ be the number of memory steps and $N:\R^{n_M+1}\to \R$ be DNN
operator that defines a DNN predictive model
\be \label{DVn}
v_{n+1} = v_n + N(V_{n}), \qquad n\geq n_M,
\ee
where
 $
V_{n} = (v_n; \cdots, v_{n-n_M}).
$
Let the initial conditions be set as the exact solution, $V_{n_M} =
U_{n_M} = (u_{n_M},\cdots, u_0)$.
Let the local truncation error of this DNN model, $\tau_{n+1}$, be defined via the
exact solution, i.e.,
\be \label{DUn}
u_{n+1} = u_n + N(U_{n})+ \Dt\cdot \tau_{n+1}.
\ee

\begin{theorem}
Consider the DNN model \eqref{DVn} with initial condition $v_n=u_n$,
$n=0,\dots, n_M$, where $n_M\geq 0$ is the number of memory
steps. For a finite time horizon $0\leq t_n\leq t_{N_T}$ where $N_T$ is finite,
assume its local truncation error is bounded and let
\be
\tau_{max} = \max_{0\leq n\leq N_T} \left\|\tau_n\right\|_\omega.
\ee
Assume that the DNN operator $N$ is Lipschitz continuous with constant
$\Lambda$ independent of $\Dt$ and $t_n\in [0,t_{N_T}]$, that is,
\be
\left\| N(V_{n}) - N(U_{n}) \right\|_\omega \leq \Lambda
\sum_{k=0}^{n_M} \left\| v_{n-k} - u_{n-k} \right\|_\omega, \quad
n\geq n_M.
\ee
Then,
\be \label{conv}
\left\| v_{n} - u_{n} \right\|_\omega\leq \tau_{max} n\Dt\cdot e^{n(n_M+1)\Lambda},
\qquad n_M\leq n \leq N_T.
\ee
\end{theorem}
\begin{proof}
  By setting $e_{n} = u_n - v_n$ and subtracting \eqref{DVn} from
  \eqref{DUn}, we obtain
  $$
  e_{n+1} = e_n + N(U_{n}) - N(V_{n}) + \Dt\cdot \tau_{n+1}, \qquad
  n_M\leq n < N_T-1.
  $$
  Summing over $n$ gives, for $j=n_M+1, \dots, N_T$,
  $$
  e_j = e_{n_M} + \Dt \sum_{n=n_M}^{j-1} \tau_{n+1} +
  \sum_{n=n_M}^{j-1} N(U_{n}) - N(V_{n}).
  $$
  Using the fact that $e_{n_M} = 0$, we obtain
  \be
  \begin{split}
  \left\|e_j\right\|_\omega &\leq \Dt \sum_{n=n_M}^{j-1} \left\|\tau_{n+1}\right\|_\omega +
  \sum_{n=n_M}^{j-1} \left\|N(U_{n}) - N(V_{n})\right\|_\omega \\
  & \leq \tau_{max} (j-n_M) \Dt  + \Lambda \sum_{n=n_M}^{j-1}  \sum_{k=0}^{n_M}
  \left\|e_{n-k}\right\|_\omega \\
  & \leq \tau_{max} j \Dt  + \Lambda  (n_M+1) \sum_{n=0}^{j-1}  
  \|e_{n}\|_\omega.
  \end{split}
  \ee
  By applying the discrete Gronwall inequality, we obtain \eqref{conv}.
 \end{proof}

 The numerical error of any DNN models is thus
 proportional to its local truncation error. 
 Since the ensemble averaged model $N^{(A)}$ produces smaller local
 truncation errors at each time step \eqref{better_lte}, it
 subsequently produces smaller numerical error during prediction than
 any of the individually trained DNN models. That is,
 \be
 \left\|v_n^{(A)} - u_n\right\|_\omega \leq \left\|v_n^{(i)} - u_n
 \right\|_\omega,
 \qquad \forall i=1,\dots, K, \quad 0\leq n\leq N_T,
 \ee
where $v^{(i)}_n$, $v_n^{(A)}$ are the scalar versions of the
individual models \eqref{Vi} and ensemble averaged model \eqref{VA}, respectively.

%% file: Examples.tex
\section{Computational Studies} \label{sec:examples}

In this section, we present numerical examples to demonstrate the properties
of the proposed ensemble averaging approach. Throughout these results, we use the terminology of \eqref{mse1} and \eqref{mse2} and compute the bias and variance components of the mean squared local truncation error in order to evaluate the accuracy gained by ensemble averaging.

\input ODE

\input Chaotic

\input PDE

%% file: ODE.tex
\subsection{Example 1: Nonlinear Pendulum System}

We first consider the following damped pendulum problem,
\[ \begin{cases} 
      \dot{x}_1 = x_2,\\
      \dot{x}_2 = -\alpha x_2-\beta \sin x_1,
   \end{cases}
\]
where $\alpha = 0.05$ and $\beta=8.91$. Training data are collected by sampling initial conditions uniformly over the computational domain of $[-\pi,\pi]\times[-2\pi,2\pi]$. A standard ResNet with 2 hidden layers with 40 nodes each is used, as in \cite{qin2018data}. The mean squared loss function is minimized using SGD with a constant learning rate of $10^{-3}$.

In all, 1000 individual DNN models are trained for 40, 80, 200, 1000, and 2000 epochs, respectively. As a reference, prediction with each individual model is carried out for one time step using the new initial condition $\mathbf{x}(0) = (-1.193, -3.876)^T$. Tables \ref{table:bias_x1}, \ref{table:bias_x2}, \ref{table:var_x1}, \ref{table:var_x2} show the bias and variance of the local truncation errors for $x_1$ and $x_2$ for the individual models in the $K=1$ (i.e. no ensemble) columns.

Next, 500 ensembles of size $K$ are chosen via sampling with replacement from the 1000 individual DNN models, and one-step ensemble averaged prediction is carried out using the same initial condition. Bias and variance are retrieved to summarize the distribution of the local truncation errors. Tables \ref{table:bias_x1} and \ref{table:bias_x2} show the bias of the error in $x_1$ and $x_2$, respectively, while Tables \ref{table:var_x1} and \ref{table:var_x2} show the variance. First, we notice from comparing Tables \ref{table:bias_x1} and \ref{table:bias_x2} that the bias is essentially unaffected by ensemble averaging, and is much more dependent on the number of training epochs. This is consistent with the idea that bias is lowered only by changing the network architecture, training data, training algorithm, or algorithm settings. In comparing the variances within each row of Tables \ref{table:var_x1} and \ref{table:var_x2}, we notice that the variance in local truncation error resulting from an ensemble model of size $K$ is roughly $1/K$ of that of the individual models $(K=1)$ regardless of the number of training epochs, which is consistent with the mathematical properties of ensemble averaging derived above.

We also note an advantageous result when comparing variances between rows in the variance tables. The computational complexity of network training, i.e. the time it takes to train a network, is proportional to the total number of training epochs. Holding this variable equal, we see, e.g. in Table \ref{table:var_x1}, that training an ensemble model of size 50 for 40 epochs (2000 total epochs) achieves a lower variance compared with training an individual model for 2000 epochs. This observation holds throughout Tables \ref{table:var_x1} and \ref{table:var_x2} when holding the total number of training epochs equal.

\begin{table}[h]
\begin{center}
\begin{tabular}{ |c|c|c|c|c| }
 \hline
 \diagbox{epochs}{$K$} & 1 & 5 & 10 & 50 \\ 
 \hline
 40 & 0.0285 & 0.0278 & 0.0284 & 0.0285 \\ 
 80 & 0.0222 & 0.0225 & 0.0225 & 0.0220 \\ 
 200 & 0.0133 & 0.0139 & 0.0133 & 0.0134 \\ 
 1000 & 0.0041& 0.0040 & 0.0040 & 0.0041 \\ 
 2000 & 0.0024 & 0.0024 & 0.0025 & 0.0024 \\ 
 \hline
\end{tabular}
\end{center}
\caption{Ex. 1: Local truncation error bias in $x_1$ for size $K$ ensemble models.}
\label{table:bias_x1}
\end{table}

\begin{table}[h]
\begin{center}
\begin{tabular}{ |c|c|c|c|c|c|c| }
 \hline
 \diagbox{epochs}{$K$} & 1 & 5 & 10 & 50 \\ 
 \hline
 40 & 0.0279 & 0.0291 & 0.0283 &0.0284 \\ 
 80 & 0.0175 & 0.0180 & 0.0179 & 0.0175 \\ 
 200 & 0.0080 & 0.0088 & 0.0080 & 0.0080 \\ 
 1000 & -0.0026 & -0.0027 & -0.0025 & -0.0025\\ 
 2000 & -0.0032 & -0.0028 & -0.0031 & -0.0032 \\ 
 \hline
\end{tabular}
\end{center}
\caption{Ex. 1: Local truncation error bias in $x_2$ for size $K$ ensemble models.}
\label{table:bias_x2}
\end{table}

\begin{table}[h]
\begin{center}
\begin{tabular}{ |c|c|c|c|c| }
 \hline
 \diagbox{epochs}{$K$} & 1 & 5 & 10 & 50 \\ 
 \hline
 40 & $1.0677\cdot10^{-3}$ & $2.2965\cdot10^{-4}$ & $1.0829\cdot10^{-4}$ & $2.3549\cdot10^{-5}$ \\ 
 80 & $6.6165\cdot10^{-4}$ & $1.3376\cdot10^{-4}$ & $6.4065\cdot10^{-5}$ & $1.3708\cdot10^{-5}$ \\
 200 & $3.7591\cdot10^{-4}$ & $7.2155\cdot10^{-5}$ & $3.6398\cdot10^{-5}$ &  $6.6175\cdot10^{-6}$ \\
 1000 & $1.0549\cdot10^{-4}$ & $2.0761\cdot10^{-5}$ & $9.8641\cdot10^{-6}$ &  $2.0956\cdot10^{-6}$ \\
 2000 & $5.4041\cdot10^{-5}$ & $1.0874\cdot10^{-5}$ & $5.1406\cdot10^{-6}$ & $1.0094\cdot10^{-6}$ \\
 \hline
\end{tabular}
\end{center}
\caption{Ex. 1: Local truncation error variance in $x_1$ for size $K$ ensemble models.}
\label{table:var_x1}
\end{table}

\begin{table}[h]
\begin{center}
\begin{tabular}{ |c|c|c|c|c|c|c| }
 \hline
 \diagbox{epochs}{$K$} & 1 & 5 & 10 & 50 \\ 
 \hline
 40 & $1.5312\cdot10^{-3}$ & $2.8850\cdot10^{-4}$ & $1.4608\cdot10^{-4}$ & $3.3492\cdot10^{-5}$ \\ 
 80 & $1.1069\cdot10^{-3}$ & $2.0120\cdot10^{-4}$ & $1.2302\cdot10^{-4}$ & $2.3370\cdot10^{-5}$ \\
 200 & $5.7697\cdot10^{-4}$ & $1.1404\cdot10^{-4}$ & $5.9217\cdot10^{-5}$ & $1.1844\cdot10^{-5}$ \\
 1000 & $1.4467\cdot10^{-4}$ & $2.9798\cdot10^{-5}$ & $1.6042\cdot10^{-5}$ &  $3.1000\cdot10^{-6}$ \\
 2000 & $7.5955\cdot10^{-5}$ & $1.5841\cdot10^{-5}$ & $7.4069\cdot10^{-6}$ & $1.5483\cdot10^{-6}$ \\
 \hline
\end{tabular}
\end{center}
\caption{Ex. 1: Local truncation error variance in $x_2$ for size $K$ ensemble models.}
\label{table:var_x2}
\end{table}

%% file: Chaotic.tex
\subsection{Example 2: Chaotic System with Missing Variables}

Next we consider a chaotic system that is only partially observed. That is, while the full system is given by
\[ \begin{cases} 
      \dot{x}_1 = -x_2-x_3,\\
      \dot{x}_2 = x_1+\frac15 x_2,\\
      \dot{x}_3 = \frac15 + y - 5x_3,\\
      \dot{y} = -\frac{y}{\epsilon} + \frac{x_1x_2}{\epsilon},
   \end{cases}
\]
with $\epsilon=0.01$, only $\mathbf{z} = (x_1,x_2,x_3)^T$ are observed. Training data are collected by sampling uniformly over the computational domain of $[-7.5,10]\times[-10,7.5]\times[0,18]\times[-1,100]$. A ResNet structure with 3 hidden layers with 30 nodes each and memory length $n_M=60$ is used, as in \cite{FuChangXiu_JMLMC20}. The mean squared loss function is minimized using Adam with a constant learning rate of $10^{-3}$.

In all, 1000 individual models are trained for 100 epochs each. As a reference, prediction with each individual model is carried out for one time step using the new initial condition $\mathbf{z}_0=(-3.7634,7.1463,13.1806)^T$ (along with $n_M$ memory) in order to look at the distribution of the local truncation error. Table \ref{table:ex2_chaotic} show the biases and variances of the local truncation errors for the individual models in the $K=1$ (i.e. no ensemble) column.

Next, 500 ensemble models of size $K=50$ are chosen via sampling with replacement from the 1000 individual DNN models, and one-step ensemble averaged prediction is carried out using the same initial condition. As in Ex. 1, Table \ref{table:ex2_chaotic} shows that the biases are not affected by ensemble averaging, while the variances of the $K=50$ ensemble models are roughly $1/K$ of that of the individual models. In addition, Table \ref{table:ex2_chaotic} shows the mean squared local truncation error (MSE) in each state variable, computed as in \eqref{mse1} and \eqref{mse2}. We see that since bias is squared in the formula, that variance dominates the MSE and so this total numerical error also exhibits a roughly $1/K$ scaling.

\begin{figure}[htbp]
	\begin{center}
		\includegraphics[width=\textwidth]{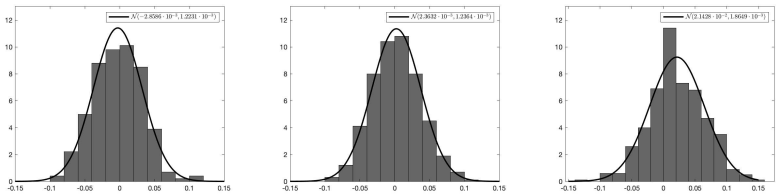}
		\caption{Ex. 2: Histograms of local truncation errors in (l-r) $x_1$, $x_2$, and $x_3$ for 500 ensemble models of size $K=50$, along with Gaussian probability density functions with corresponding statistics from Table \ref{table:ex2_chaotic}.}
		\label{fig:chaotic_histogram}
	\end{center}
\end{figure}

Figure \ref{fig:chaotic_histogram} shows the distributions of the local truncation errors in each of the observed variables in histograms compared with a normal distribution with corresponding biases and variances from Table \ref{table:ex2_chaotic}. For all three observed variables, the local truncation errors fail to reject a chi-square goodness-of-fit test with null hypothesis that the errors come from a Gaussian distribution with the corresponding bias and variance at $0.05$ significance level. This demonstrates the scalar version of \eqref{local_WA} in this case.

\begin{table}[h]
\begin{center}
\begin{tabular}{ |c|c|c| }
 \hline
$K$ & 1 & 50 \\ 
 \hline
$\text{Bias}({x_1})$ & $-2.8666\cdot10^{-3}$ & $-2.8586\cdot10^{-3}$ \\
$\text{Bias}({x_2})$ & $-1.1997\cdot10^{-2}$ & $2.3632\cdot10^{-3}$ \\ 
$\text{Bias}({x_3})$ & $2.2516\cdot10^{-2}$ & $2.1428\cdot10^{-2}$ \\ 
\hline
$\text{Var}({x_1})$ & $6.4272\cdot10^{-2}$ & $1.2231\cdot10^{-3}$ \\
$\text{Var}({x_2})$ & $5.9915\cdot10^{-2}$ & $1.2364\cdot10^{-3}$ \\ 
$\text{Var}({x_3})$ & $8.8858\cdot10^{-2}$ & $1.8649\cdot10^{-3}$ \\ 
\hline
$\text{MSE}(x_1)$ & $6.4280\cdot10^{-2}$ & $1.2312\cdot10^{-3}$ \\ 
$\text{MSE}(x_2)$ & $6.0059\cdot10^{-2}$ & $1.2420\cdot10^{-3}$ \\ 
$\text{MSE}(x_3)$ & $8.9365\cdot10^{-2}$ & $2.3241\cdot10^{-3}$ \\ 
 \hline
\end{tabular}
\end{center}
\caption{Ex. 2: Local truncation error statistics for ensemble models of size $K$.}
\label{table:ex2_chaotic}
\end{table}




%% file: PDE.tex
\subsection{Example 3: Partial Differential Equation System} \label{sec:pdesystem}



Finally, we discuss the proposed ensemble averaged DNN learning for systems of PDEs. As in \cite{chen2022deep}, we consider the linear wave system
\begin{align}\label{eq:vecsystem}
\mathbf{u}_t = \A\mathbf{u}_x, { \quad x \in [0,2\pi],}
\end{align}
where
\begin{align*}
\mathbf{u} = \begin{bmatrix} u_1(x,t) \\ u_2(x,t)\end{bmatrix} \quad\text{and}\quad \A=\begin{bmatrix} 0 & 1 \\ 1 & 0 \end{bmatrix},
\end{align*}
with $2\pi$-period boundary condition. (Note that even though its
form is simple, linear wave equations like this are exceptionally
difficult to solve accurately using numerical methods.)
Training data are collected over a uniform grid with $N=50$ points by computing
the exact solution of the
true system under randomized initial conditions in the form of the
Fourier series
\begin{equation} \label{Fourier}
	u(x,0) = a_0+ \sum_{n=1}^{N_c} \left ( a_n\cos(nx) +
          b_n\sin(nx) \right ).
\end{equation}
In particular, the initial conditions for
both components $u_1$ and $u_2$ are generated by sampling
\eqref{Fourier} with $a_0\sim U[-\frac{1}{2},\frac{1}{2}]$, and
$a_n ,b_n\sim U\left[-\frac{1}{n} ,\frac{1}{n}\right]$ for $1\le n \le N_c=10$. 
The DNN structure consists of an input layer with $100$ neurons, to
account for the two components $u_1$ and $u_2$, each of which has $50$
spatial nodes. The disassembly block has dimension $100\times 1 \times 5$ and the
assembly layer has dimension $1\times1\times 5$, as in \cite{chen2022deep}. 
The mean squared loss function is minimized using Adam with a constant learning rate of $10^{-3}$.

In all, 1000 individual models are trained for 40 epochs each. 
As a reference, prediction with each individual model is carried out for one time step using the new initial condition
\begin{align*}
\mathbf{u}(x,0) &= \begin{bmatrix} \exp(\sin(x)) \\ \exp(\cos(x)) \end{bmatrix}.
\end{align*}
Since in this case the large spatial grid results in 100 state variables, the 2-norm of the local truncation errors in $\u$ are computed.
Table \ref{table:ex3_pde} shows the bias and variance of these normed local truncation errors for the individual models in the $K=1$ (i.e. no ensemble) columns.

Next, 500 ensemble models each of sizes $K=2,5,10$ are chosen via sampling with replacement from the 1000 individual DNN models, and one-step ensemble averaged prediction is carried out for one time step using the same initial condition. Bias and variance are retrieved to summarize the distribution of the 2-norm of the local truncation errors. Table \ref{table:ex3_pde} shows consistency with the previous examples in that the biases are similar while the variances differ roughly by a factor of $1/K$. The mean squared normed local truncation error (MSE) is also computed. In this case, unlike Ex. 2, the squared bias actually dominates the MSE, yet this total numerical error still exhibits a roughly $1/K$ scaling.


\begin{table}[h]
\begin{center}
\begin{tabular}{ |c|c|c|c|c|c|c| }
 \hline
$K$ & 1 & 2 & 5 & 10 \\ 
 \hline
$\textrm{Bias}(\|\u\|_2)$ & $6.2115\cdot10^{-3}$ & $4.5229\cdot10^{-3}$ & $2.8915\cdot10^{-3}$ & $2.1219\cdot10^{-3}$ \\ 
\hline
$\text{Var}(\|\u\|_2)$ & $5.6610\cdot 10^{-6}$ & $1.6189\cdot10^{-6}$ & $3.9791\cdot10^{-7}$ & $1.4576\cdot10^{-7}$ \\
\hline
$\text{MSE}(\|\u\|_2)$ & $4.4244\cdot 10^{-5}$ & $2.2075\cdot10^{-5}$ & $8.7588\cdot10^{-6}$ & $4.6482\cdot10^{-6}$ \\
 \hline
\end{tabular}
\end{center}
\caption{Ex. 3: Normed local truncation error statistics for size $K$ ensemble models.}
\label{table:ex3_pde}
\end{table}

%% file: Conclusion.tex
\section{Conclusion} \label{sec:conclusions}

We present a general ensemble averaging method for robust DNN learning of unknown systems in which one-step predictions from multiple models are averaged and used as the input for the next time step prediction. Mathematical foundation for undertaking such a scheme is provided that establishes a probability distribution and an error bound for the local truncation error, demonstrating that the variance of the local truncation error for a size $K$ ensemble averaged model is roughly $1/K$ that of an individual model. A set of three examples are presented which confirms the properties predicted by the mathematical foundation and shows that the proposed approach is able to robustly model a variety of standard differential equation problems from data.